# Interpretability in Convolutional Neural Networks for Building Damage Classification in Satellite Imagery

Thomas Y. Chen
The Academy for Mathematics, Science, and Engineering
thomaschen7@acm.org

## Abstract

Natural disasters ravage the world's cities, valleys, and shores on a regular basis. Deploying precise and efficient computational mechanisms for assessing infrastructure damage is essential to channel resources and minimize the loss of life. Using a dataset that includes labeled pre- and post- disaster satellite imagery, we take a machine learning-based remote sensing approach and train multiple convolutional neural networks (CNNs) to assess building damage on a per-building basis. We present a novel methodology of interpretable deep learning that seeks to explicitly investigate the most useful modalities of information in the training data to create an accurate classification model. We also investigate which loss functions best optimize these models. Our findings include that ordinal-cross entropy loss is the most optimal criterion for optimization to use and that including the type of disaster that caused the damage in combination with pre- and post-disaster training data most accurately predicts the level of damage caused. Further, we make progress in the qualitative representation of which parts of the images that the model is using to predict damage levels, through gradient-weighted class activation mapping (Grad-CAM). Our research seeks to computationally contribute to aiding in this ongoing and growing humanitarian crisis, heightened by anthropogenic climate change.

## 1 Introduction

Natural disasters devastate numerous vulnerable communities and countries annually. They are responsible for the deaths of 60,000 people a year worldwide, on average [18]. The timely allocation of resources in the event of these occurrences is crucial to saving lives. Additionally, natural disasters cause varying levels of damage to buildings. The havoc wreaked by them causes widespread infrastructure damage, in some cases leading to a "cascade effect" [15]. The resulting economic impact is significant. For example, since 1980, the United States has sustained 273 weather and climate disasters that have caused damages exceeding 1 billion US dollars (USD), totaling 1.79 trillion USD [6]. The frequency and severity of these disasters is projected to only continue to increase, exacerbated by increasing global surface temperatures as a result of climate change [22]. The catastrophic impact of natural disasters and their increasing prevalence motivates the problem addressed in this work. In order to prepare for and recover from these events, robust emergency response plans must be in place. This requires quickly and accurately analyzed data from the disaster site. Because it is relatively difficult to obtain damage assessment and other details from on the ground in a timely manner, satellite imagery has gained popularity in being used to analyze these types of situations. Deep neural networks (DNNs) have been used to locate and classify building damage within satellite imagery [12, 11, 25, 9]. However, the current literature is limited in the interpretability of what exactly these neural networks are learning and what is most useful in assessing building damage.

To address this problem, we present a novel analysis of the most important information that a deep learning model needs to assess building damage. We use a convolutional neural network (CNN) architecture called a residual neural network (ResNet), pre-trained on Imagenet data [13]. In our approach, we train multiple CNNs on xBD satelite imagery data [10], with different modalities of input, as well as using different loss functions, and compare accuracy on the validation set. We aim to explicitly provide insight into the most effective ways to train models to classify levels of building damage, maximizing the efficiency of the emergency response after a natural disaster, which has the potential to save lives and reduce economic strain.

## 2 Related Works

Satellite imagery is useful in a wide range of application areas, including for assessment of marine ecology [21], weather forecasting [4], and even in studying and predicting the spread of infectious disease [19]. Lately, there has been increased interest in using satellite imagery for humanitarian purposes such as responding to natural disasters [17]. Satellite imagery also provides insights for agriculture [26] and urban road damage [16]. More generally, change detection, which is the process of identifying differences in the state of an object by observing it at different times, can be used with satellite imagery in a variety of contexts. There are a few primary categories in which change detection approaches fall: algebra-based, transform-based, and classification-based [1]. Change detection has been employed on satellite imagery to study deforestation [23], urban growth [14], and more. One specific area that has garnered significant attention in computer vision and satellite imagery is building damage assessment. Recent works have studied semantic building segmentation [12, 11] and cross-region transfer learning to avoid overfitting [25]. Furthermore, [9] presents a semi-supervised approach. xView2 recently introduces a dataset, xBD, discussed more in detail later in this work [10]. Many teams [24] competed in the xView2 data competition and improved upon the baseline model provided [10]. In our work, we focus on building damage assessment via image classification and change detection. We specifically hone in on what information is most useful in accurate classifications of building damage and analyze which loss functions are most fit for training our models and yield the most efficacious results. Our primary contribution is to improve upon the interpretability of machine learning models of prior works and existing literature in this area by explicitly examining per-building classification prediction accuracy with different combinations of inputted information and loss functions.

## 3 Methods

### 3.1 Dataset Details

For this work, we utilize the xBD dataset [10], which covers a wide range of disasters in fifteen countries around the world, from Guatemala to Portugal to Indonesia (over 850,736 building polygons totaling an area of 45,361 square kilometers). One of xBD's main purposes is to demonstrate changes between pre-disaster and post-disaster satellite imagery to aid in detecting the damage caused. Therefore, each post-disaster building is labeled as one of the following: "unclassified," "no damage," "minor damage," "major damage," or "destroyed." (We discard the "unclassified" buildings). The classification benchmark utilized is called the Joint Damage Scale (JDS). We use the xBD dataset because it incorporates a variety of disaster types, building types, and geographical locations. This allows for diversity in training the model. For example, the wide variety of geographical locations is important for cross-region generalization. Additionally, the high resolution imagery allows for detailed change detection between pre-disaster and post-disaster images. These factors currently make xBD the leading dataset for building damage detection using labeled satellite imagery [10]. Previous satellite imagery datasets were not as comprehensive, and, for example, had only covered singular disaster types or did not have uniform building damage assessment criteria like the JDS used in xBD [8, 2, 7].

### 3.2 Data Preprocessing

The dataset consists of 1024 by 1024 pixel satellite images. In order to zero in on the changes, we begin by collecting bounding boxes of the buildings in each image from the segmentation ground truth masks (building polygons) provided. We discard buildings that have a bounding box size of less than 2,000 pixels, as they are too small and blurred to be valuable training data, possibly hindering the model from achieving accurate results. We also discard any buildings with the classification ground truth label of "unclassified," because this information is not useful for our purposes. In order to maintain an equal distribution over JDS classification (damage level) in our training and validation sets so that we can properly assess model accuracy, we provide for an equal number of buildings of the categories "destroyed," "major damage," "minor damage," and "no damage" in each set, while still maintaining a 0.8:0.2 ratio between train and validation. The xBD dataset is deliberately created with a disproportionately large volume of buildings with no damage [10], but training on such a lopsided data distribution would yield artificially high accuracy numbers and not yield valuable results.

### 3.3 Baseline Model

We train a baseline classification model to classify buildings by damage level, as defined by the Joint Damage Scale. The model input is only the post-disaster image. Notably, our baseline model does not use change detection. Because the data is labeled, this is a supervised approach. The model architecture is ResNet18, an 18 layer CNN, and was pre-trained on ImageNet data [5]. This baseline model uses the cross-entropy loss function, which is defined as

$$-\sum_{c=1}^{4} y_{o,c} \log(p_{o,c}),$$

where $y_{o,c}$ is a binary indicator (either 0 or 1) of whether $c$, as a label, correctly classifies observation $o$, and $p_{o,c}$ is the predicted probability that observation $o$ is of the class $c$. Cross-entropy loss is defined, in other terms, as the negative sum of the expression $y_{o,c} \log(p_{o,c})$ across all 4 possible classes $c$: no damage, minor damage, major damage, and destroyed. The network is trained on 12,800 buildings crops with a batch size of 32. The Adam optimizer with a learning rate of 0.001 is used. The model trained for 100 epochs on NVIDIA Tesla K80 GPUs.

### 3.4 Improvements

We train other models that improve upon the performance of the baseline model. We introduce other model inputs, namely the pre-disaster image (in combination with the post-disaster image) and the type of disaster (e.g. volcano, wind, etc.) that caused the building damage. To train a model that takes in both pre-disaster images and their corresponding post-disaster images, we concatenate the RGB channels of the two and use that as input. To train a model that takes in the pre-disaster image, post-disaster image, and disaster type, we do the same, but also concatenate a one-hot encoded representation of the disaster type in one of the later layers of the CNN.

Furthermore, we experiment with other loss functions, namely mean squared error loss and ordinal cross-entropy loss to train these models. We define mean squared error as

$$\frac{1}{b} \sum_{i=1}^{b} (y - \hat{y})^2,$$

where $b$ is the batch size, $y$ is the ground truth (a class from 0 to 3 representing each damage level), and $\hat{y}$ is the prediction. Ordinal cross-entropy loss differs from cross-entropy loss in that it takes into account the distance between the ground truth and the predicted class (hence "ordinal"). Since the building damage classification problem involves different and increasing levels of damage from no damage to destruction, this function is useful to distinguish between different categories. To implement ordinal cross-entropy loss as the loss function, we treat it as generic multi-class classification and encode the classes no damage, minor damage, major damage, and destroyed as [0, 0, 0], [1, 0, 0], [1, 1, 0], and [1, 1, 1], respectively [3]. The other aspects of the training process (optimizer, learning rate, number of epochs, etc.) remain the same. These improved models contribute to our understanding of what information leads to the most accurate prediction results for building damage assessment.

## 4  Results and Discussion

In Table 1, we present model accuracy on the validation set across nine different models, which are differentiated by three different input combinations and three different loss functions. The baseline model, which is trained with post-disaster data only and the cross-entropy loss function, has an accuracy of 59.5%, as shown. It is important to note that all models were trained and validated on data that is evenly split between building crops of each class (no damage, minor damage, major damage, and destroyed), so a purely blind guessing model would achieve approximately 25% accuracy.

Table 1: Comparison of Validation Accuracy on 9 Different Models

| | Model Accuracy on Validation Set with Chosen Loss (100 epochs) | | |
|---|---|---|---|
| | **Loss Function** | | |
| **Model Input** | Mean Squared Error | Cross-Entropy Loss | Ordinal Cross-Entropy Loss |
| Post-Disaster Image Only | 45.3% | 59.5% | 64.2% |
| Pre-Disaster, Post-Disaster Images | 50.2% | 68.3% | 71.2% |
| Pre-Disaster, Post-Disaster Images, Disaster Type | 49.7% | 72.7% | 74.6% |

When the model is trained and validated on both pre-disaster and post-disaster building imagery as opposed to solely the post-disaster data, we see an 8.8% increase in accuracy on the validation set in comparison to the baseline model, while keeping the loss function constant. Adding the disaster type as a third type of input subsequently increases accuracy by another 4.4%. Reverting to the baseline model, changing the loss function utilized to ordinal cross-entropy loss instead of cross-entropy loss, we see a 4.7% accuracy jump on the validation set. Sticking with ordinal cross-entropy loss, adding the pre-disaster image as a mode of input increases accuracy by another 7.0%, while adding the disaster type as another mode of input increases accuracy by an additional 3.4%.

Our results largely conform to our hypotheses. Firstly, accuracy on the validation set improves when more modes of useful information are inputted into the model (accuracy generally increases moving down the rows of Table 1). This is reasonable because the more relevant information that the model has to work with, the more accurate predictions it should make. A large part of our research was addressing which types of input aid the convolutional neural networks in making accurate predictions. From the results generated, it seems that having the aspect of change detection (when the pre-disaster image is concatenated with the post-disaster image and inputted) is useful, along with the type of disaster. We also note that models using ordinal cross-entropy loss as their criterion for optimization perform the most accurately. This is also reasonable because, as previously mentioned, ordinal cross-entropy loss is most specifically applicable for a classification problem that involves an ordinal scale (in this case, the JDS), as opposed to categories with no intrinsic ordering. Mean squared error (MSE), not surprisingly, demonstrated to be the least effective loss function to use for training. This is justifiable because MSE is primarily used in regression problems, not classification problems. We find that cross-entropy loss models fall somewhere in between.

However, we note that none of the accuracy numbers are necessarily optimal. This can be explained by the fact that the differences between categories, particularly between minor-damage and major-damage, are often difficult to discern, for both humans and AI. This is certainly a challenge with non-binary classification tasks with building damage that has been acknowledged by many, including Gupta et. al [10]. Additionally, more thorough dataset cleaning may yield marginally more accurate results. These results contribute to the research area of building damage detection by addressing the limited interpretability of deep learning algorithms in the current literature in regards to what types of information are most useful to building damage classification models as well as which loss functions are the best criteria for optimization.

## 5 Conclusion

The main insights that can be drawn from our work include using individualized building crops instead of semantic segmentation to train models and performing experiments with various combinations of model inputs and loss functions to explicitly examine their differences. Our work's main contribution to the field is presenting a novel, more interpretability-oriented, analysis of how to classify building damage most accurately and effectively in the event of a natural disaster. Practically, our work and similar works in the field advance methods for more robust emergency responses and more efficient allocation of resources, which saves lives and property. This research is especially important now, when climate change is ramping up the frequency and intensity of these devastating events.

Building upon our work with improving interpretability, future work includes investigations of the prediction performance of deep learning models with other types of input added, such as neighboring building ground truths. Additionally, other ways to combine information such as pre-disaster and post-disaster images (instead of a simple concatenation like we did here) with the goal of improving interpretability should yield valuable results.

## 6 Acknowledgements

The author thanks Ethan Weber (Massachusetts Institute of Technology) for his mentorship during the ideation, experimental design, and overall research processes. The author also thanks Climate Change AI (CCAI) and the NeurIPS 2020 CCAI Workshop Organizing and Program Committees.

# A  Qualitative Results

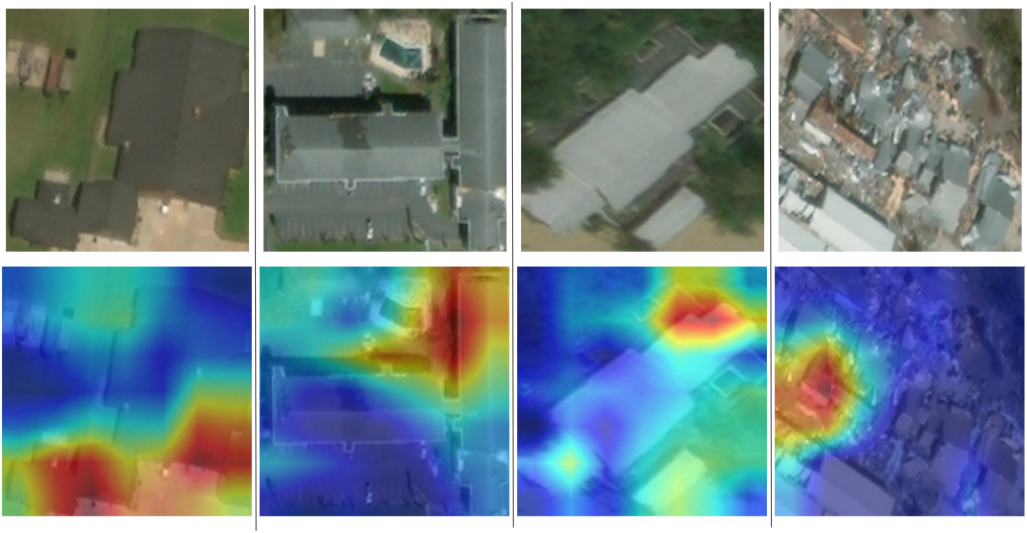

Figure 1: Gradient class activation maps [20] depict which parts of the building crop lead the baseline model to predict a certain classification. On the first row are the original images (crops) and on the second row are the corresponding gradient class activation maps. The images included consist of solely post-disaster images. From left to right: (1) A building with label "no damage," after flooding in the Midwestern United States, (2) A building with label "minor damage," after Hurricane Michael, (3) A building with label "major damage," after Hurricane Harvey, and (4) A building with label "destroyed," after Hurricane Michael.

# B Additional Dataset Details

## B.1 Building Size

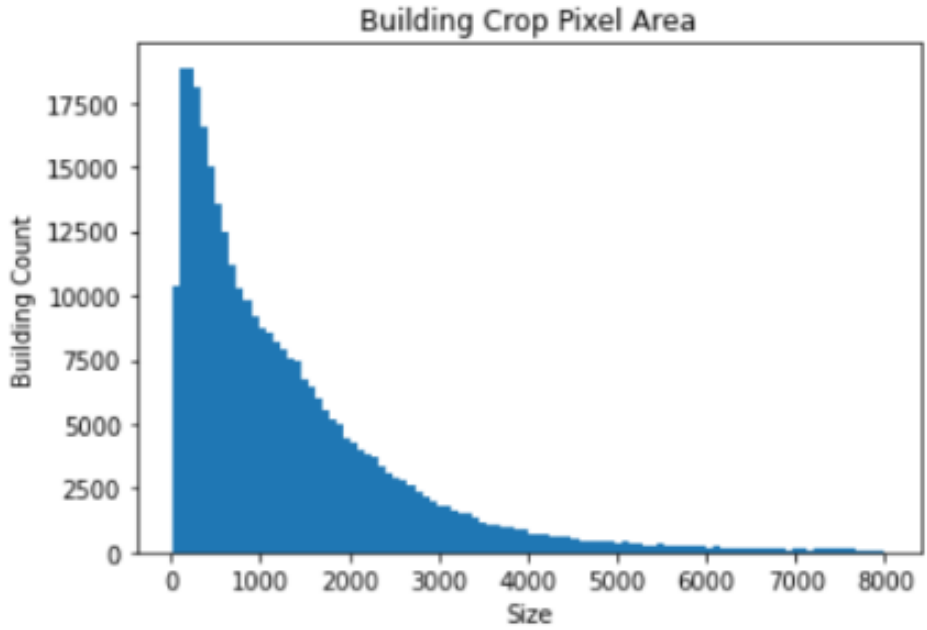

Figure 2: A histogram representing the distribution of building bounding box areas in the xBD dataset. Outliers (bounding boxes with a pixel area greater than 8000) have been removed from the graph.

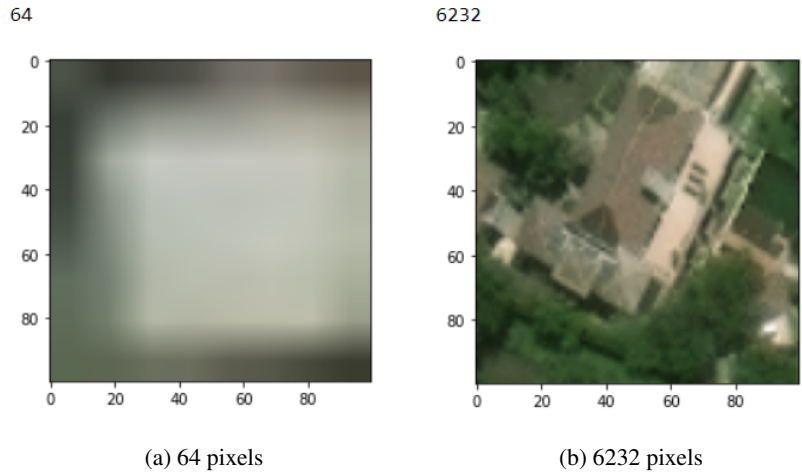

(a) 64 pixels

(b) 6232 pixels

Figure 3: A comparison of two vastly differently sized buildings in the dataset. Both are resized to 100 by 100 pixels for viewing. The blurred building of size 64 pixels would not be a useful data point for any deep learning model to learn from and yield accurate results. Instead, buildings like this are noisy and are discarded in our methodology.